\documentclass[conference]{IEEEtran}
\makeatletter
\def\ps@headings{%
\def\@oddhead{\mbox{}\scriptsize\rightmark \hfil \thepage}%
\def\@evenhead{\scriptsize\thepage \hfil \leftmark\mbox{}}%
\def\@oddfoot{}%
\def\@evenfoot{}}
\makeatother
\usepackage[justification=centering]{caption}
\usepackage{url}
\usepackage{booktabs}
\usepackage{graphicx,subfigure}
\usepackage{epstopdf}
\usepackage{amsmath}
\usepackage{algorithm}
\usepackage{algpseudocode}
\usepackage{amsmath}
\usepackage{amssymb}
\usepackage{amsthm}
\usepackage{epsfig}

\usepackage{amsfonts}
\usepackage{multirow}
\usepackage{color}
\usepackage{array}
\usepackage{listings}
\usepackage{hyperref}
\usepackage[underline=true]{pgf-umlsd}

\usepackage{setspace}

\begin{document}

\title{Text Summarization Using Large Language Models: A Comparative Study of MPT-7b-instruct, Falcon-7b-instruct, and OpenAI Chat-GPT Models}

\author{\IEEEauthorblockN{Lochan Basyal}
\IEEEauthorblockA{\textit{Mentee, KaggleX BIPOC Mentorship Program Cohort 3} \\
bashyallochan@gmail.com}
\and
\IEEEauthorblockN{Mihir Sanghvi}
\IEEEauthorblockA{\textit{Mentor, KaggleX BIPOC Mentorship Program Cohort 3} \\
mihir2891@gmail.com}
}

\maketitle

\begin{abstract}
Text summarization is a critical Natural Language Processing (NLP) task with applications ranging from information retrieval to content generation. Leveraging Large Language Models (LLMs) has shown remarkable promise in enhancing summarization techniques. This paper embarks on an exploration of text summarization with a diverse set of LLMs, including MPT-7b-instruct, falcon-7b-instruct, and OpenAI ChatGPT text-davinci-003 models. The experiment was performed with different hyperparameters and evaluated the generated summaries using widely accepted metrics such as the Bilingual Evaluation Understudy (BLEU) Score, Recall-Oriented Understudy for Gisting Evaluation (ROUGE) Score, and Bidirectional Encoder Representations from Transformers (BERT) Score. According to the experiment, text-davinci-003 outperformed the others. This investigation involved two distinct datasets: CNN Daily Mail and XSum. Its primary objective was to provide a comprehensive understanding of the performance of Large Language Models (LLMs) when applied to different datasets. The assessment of these models' effectiveness contributes valuable insights to researchers and practitioners within the NLP domain. This work serves as a resource for those interested in harnessing the potential of LLMs for text summarization and lays the foundation for the development of advanced Generative AI applications aimed at addressing a wide spectrum of business challenges.
\end{abstract}

\begin{IEEEkeywords}
Text Summarization, MPT-7b-instruct, Falcon-7b-instruct, OpenAI ChatGPT
\end{IEEEkeywords}

\section{Introduction}
In the era of Big Data, the abundance of textual information has underscored the importance of efficient text summarization techniques. Text summarization, the task of distilling long documents or articles into concise, coherent summaries while preserving the core meaning and essential information, holds immense value across various domains. From aiding in information retrieval to content generation, summarization has emerged as a pivotal component of Natural Language Processing (NLP) applications.

Recent advancements in NLP have been characterized by the rise of Large Language Models (LLMs), OpenAI ChatGPT\cite{chatgpt}, MPT-7b-instruct\cite{mpt, mpt-huggingface}, flan-t5-xl\cite{t5xl}, falcon-7b-instruct\cite{falcon, falcon40b}, and others, which have demonstrated remarkable capabilities in understanding and generating human-like text. These LLMs have opened new avenues for text summarization by providing powerful generative capabilities and the ability to adapt to diverse tasks through fine-tuning.

The 'text-davinci-003 (Legacy)'\cite{chatgpt} model represents a remarkable leap in the field of Natural Language Processing (NLP). It exhibits an unparalleled ability to handle a wide range of language tasks with exceptional precision and quality. Notably, it surpasses its predecessors, including the Curie, babbage, and Ada models, in terms of generating text of higher quality, offering longer outputs, and consistently following provided instructions. This legacy model has a token capacity of 4,097, enabling it to handle extensive text generation with ease. Moreover, 'text-davinci-003 (Legacy)' introduces innovative features, such as the capability to insert text seamlessly into generated content, thereby expanding its utility for diverse text manipulation tasks.

The 'MPT-7B-Instruct' model, as cited in\cite{mpt, mpt-huggingface}, is specifically designed for short-form instruction-following tasks, making it an ideal choice for a wide range of instruction-based applications. It is created through the fine-tuning process of a base model, MPT-7B, using a dataset sourced from the Databricks Dolly-15k and the Anthropic Helpful and Harmless (HH-RLHF) datasets. This tailored approach results in a model that excels at understanding and following instructions with precision and accuracy. The model follows a modified decoder-only transformer architecture, optimized for superior performance in instruction-following tasks.

Falcon-7B-Instruct', as cited in\cite{falcon, falcon40b}, represents a formidable 7 billion-parameter causal decoder-only model meticulously crafted by the Technology Innovation Institute (TII). This model is built upon the robust foundation of Falcon-7B and undergoes a fine-tuning process using a composite dataset sourced from both chat and instruct domains. 'Falcon-7B-Instruct' is generously made available under the Apache 2.0 license.

The focus of this paper is to delve into the world of text summarization with LLMs, offering a comprehensive exploration of their potential and limitations. Specifically, we investigate various LLMs, experiment with different hyperparameters, and evaluate the quality of summaries generated by these models. To ensure a robust evaluation, we employ well-established metrics such as BLEU Score, Rouge Score, and Bert Score.

This paper serves as a vital resource for those seeking to harness the power of LLMs for NLP applications and lays the groundwork for the development of advanced Generative AI solutions to address a wide range of business challenges. In the following sections, the paper provides detailed explanations of the text summarization methods discussed in Section II, supervised and unsupervised summarization in Section III, datasets and evaluation metrics presented in Section IV, inference with different LLMs in Section V, and offers a roadmap for future enhancements, concluding with Section VI. Lastly, the author acknowledges the support received during the research and experiments.

\section{Text Summarization Methods}
Text summarization is a fundamental task in Natural Language Processing (NLP) that aims to condense large volumes of text into shorter, coherent representations while preserving the essential information. There are primarily two approaches to text summarization: abstractive and extractive summarization.

\subsection{Abstractive Text Summarization}
Abstractive summarization involves generating a concise summary that may contain words, phrases, or sentences not present in the source text. This approach relies on understanding the context and generating human-like language to convey the central ideas. Abstractive summarization methods often use advanced language models, such as Large Language Models (LLMs), to rewrite and rephrase content in a more concise form.

\subsection{Extractive Text Summarization}
Extractive summarization, on the other hand, aims to select and extract the most important sentences or phrases directly from the source text to form the summary. It does not involve rephrasing or generating new sentences. Extractive summarization methods use various techniques, such as sentence scoring and ranking, to identify and extract the most salient content.

\section{Supervised and Unsupervised Summarization}
Text summarization techniques can be broadly categorized into two main approaches based on dataset labeling: supervised and unsupervised summarization. Each approach has its methodologies and advantages, serving different use cases and data availability scenarios.

\subsection{Supervised Summarization}
Supervised summarization is a method that relies on labeled training data, where human annotators provide summaries for a given set of source texts. Machine learning models are then trained on this data to learn the mapping between source texts and their corresponding summaries. This approach is particularly effective when high-quality, domain-specific summaries are available for training.

\subsection{Unsupervised Summarization}
Unsupervised summarization, on the other hand, does not require labeled training data. Instead, it seeks to extract the most relevant information from the source text using algorithms that consider factors like sentence importance, coherence, and redundancy. Unsupervised methods are often employed when labeled summarization datasets are scarce or costly to obtain.

\section{Datasets and Evaluation Metrics}
In our study, we conducted experiments and evaluations on two distinct datasets, CNN/Daily Mail 3.0.0\cite{cnn} and the Extreme Summarization (XSum)\cite{xsum} to assess the performance of various Large Language Models (LLMs) in the context of text summarization. These datasets serve as the foundation for our evaluation and comparison of LLM-generated summaries.

\subsection{Datasets}
\begin{itemize}
    \item \textbf{CNN/Daily Mail 3.0.0 Dataset:} The CNN/Daily Mail 3.0.0 Dataset, a valuable resource in the realm of natural language processing, comprises more than 300,000 unique news articles authored by journalists from CNN and the Daily Mail. Originally designed to facilitate machine reading and comprehension, this English-language dataset has since evolved to support both extractive and abstractive summarization tasks.
    The dataset provides three key data fields for each entry: 'id,' which contains the hexadecimal-formatted SHA1 hash of the URL from which the story was retrieved; 'article,' which contains the body of the news article itself; and 'highlights,' featuring the article's highlights as written by the original author.
    \item \textbf{XSum Dataset:} XSum dataset is a valuable resource tailored for extreme summarization tasks. It consists of news articles with three key features: the 'document,' serving as the input news article, the 'summary' providing a one-sentence summary of the article, and the 'id,' which uniquely identifies each article using the BBC ID. 
\end{itemize}

The inclusion of these diverse datasets allows us to evaluate the performance of LLMs across various content types, ensuring that our study provides a holistic view of their summarization capabilities.

\subsection{Evaluation Metrics}
To assess the quality and effectiveness of the generated summaries, we employed a set of widely accepted evaluation metrics:
\begin{itemize}
    \item \textbf{BLEU Score\cite{bleuscore}:} BLEU is a metric employed to assess the quality of machine translations. It operates by measuring the similarity between n-grams present in machine-translated sentences and those in human-translated sentences. It is generally noted that the BLEU score tends to decrease with longer sentence lengths, although variations in this trend can occur depending on the translation model in use.
    \item \textbf{ROUGE Score\cite{rougescore, metrics}:} The ROUGE Score assesses the overlap of n-grams (sequences of words) between the generated summary and reference summaries. It considers metrics such as ROUGE-N (unigrams, bigrams, etc.) and ROUGE-L (longest common subsequence) to evaluate content overlap.
    \item \textbf{BERT Score\cite{bertscore, metrics}:} The BERT Score utilizes contextual embeddings from the BERT model to measure the similarity between the generated summary and reference summaries. It is designed to capture the nuances of language and context, providing a robust evaluation metric.
\end{itemize}

By calculating these metrics for summaries generated with different LLMs, we aim to provide a comprehensive assessment of their performance, enabling researchers and practitioners to make informed decisions when choosing an LLM and fine-tuning their summarization models for specific tasks and datasets.

\section{Inference with Different LLMs}
In this section, the results of the experiments are presented, wherein a variety of Large Language Models (LLMs) were utilized to generate summaries for two distinct datasets. The LLMs employed for these experiments include falcon-7b-instruct, mpt-7b-instruct, and text-davinci-003. The primary objective is to offer a comparative analysis of their performance concerning text summarization.

\subsection{Experiment Setup}
For each LLM, experiments were conducted using a temperature value of 0.1 and a maximum token length of 100. These experiments involved summarizing 25 test samples of each dataset. The process of generating the text summary entailed the utilization of LangChain and Hugging Face pipelines for prompt engineering, ensuring precision and efficiency in the summarization process. This experiment was executed by hosting custom Google Compute Engine Virtual Machine (GCE VM) instances equipped with NVIDIA T4 Graphics Processing Units (GPUs) sourced from the Google Cloud Platform (GCP).

\subsection{Results}
The performance of different LLMs on two distinct datasets, utilizing the specified temperature value, was displayed. Metrics were computed for each LLM, offering a comprehensive perspective on their summarization capabilities, as available on the GitHub repository cited in this paper\cite{code}.

These tables, as referenced in Table~\ref{table:performance-metrics} and Table~\ref{table:performance-metrics-bleu}, present a comprehensive evaluation of various Large Language Models (LLMs) for text summarization across two distinct datasets: CNN/Daily Mail 3.0.0 and XSum. The performance of each LLM is assessed using several key metrics, including BLEU, ROUGE, and BERT.

The table highlights varying performance across LLMs and datasets. Notably, the OpenAI model, text-davinci-003, consistently exhibits strong performance, achieving high BLEU, ROUGE, and BERT Scores. This exceptional performance can be attributed to davinci being the largest and most powerful model, with 175 billion parameters and 45TB of text data. When comparing the two 7b parameter fine-tuned models, MPT-7b-instruct performed slightly better than Falcon-7b-instruct. However, their overall performance was somewhat similar. These findings underscore the significance of model architecture and size in text summarization tasks, as well as the potential of OpenAI's model for achieving state-of-the-art results in diverse NLP applications.

\begin{table*}[t!]
\centering
\begin{tabular}{ccccccc}
\toprule
\textbf{LLM Model} & \textbf{Dataset} & \textbf{Avg. Word Count} & \textbf{ROUGE-1} & \textbf{ROUGE-2} & \textbf{ROUGE-L} & \textbf{BERT Score (P/R/F1)} \\
\hline
falcon-7b-instruct &  CNN (n=25) & 784.24 & 0.226 & 0.053 & 0.197 & 0.818 / 0.860 / 0.838 \\
falcon-7b-instruct & XSum (n=25) & 410.44 & 0.139 & 0.014 & 0.113 & 0.787 / 0.863 / 0.823 \\
mpt-7b-instruct & CNN (n=25) & 784.24 & 0.236 & 0.060 & 0.213 & 0.839 / 0.864 / 0.851 \\
mpt-7b-instruct & XSum (n=25) & 410.44 & 0.159 & 0.024 & 0.133 & 0.828 / 0.871 / 0.848 \\
text-davinci-003 & CNN (n=25) & 784.24 & 0.272 & 0.096 & 0.255 & 0.854 / 0.883 / 0.868 \\
text-davinci-003 & XSum (n=25) & 410.44 & 0.206 & 0.053 & 0.173 & 0.844 / 0.893 / 0.868 \\

\hline
\end{tabular}
\caption{Performance Metrics of LLMs on "CNN/Daily Mail
3.0.0" and "XSum" Datasets}
\label{table:performance-metrics}
\end{table*}

\begin{table*}[t!]
\centering
\begin{tabular}{ccccccc}
\toprule
\textbf{LLM Model} & \textbf{Dataset} & \textbf{Avg. Word Count} & \textbf{BLEU Score}\\
\hline
falcon-7b-instruct &  CNN (n=25) & 784.24 & 9.4726138403298e-232 \\
falcon-7b-instruct & XSum (n=25) & 410.44 & 9.225829346520394e-232 \\
mpt-7b-instruct & CNN (n=25) & 784.24 & 9.35328936831654e-232 \\
mpt-7b-instruct & XSum (n=25) & 410.44 &  9.542118736121376e-232 \\
text-davinci-003 & CNN (n=25) & 784.24 &  0.4896200481408649 \\
text-davinci-003 & XSum (n=25) & 410.44 &  0.48979461356547943 \\

\hline
\end{tabular}
\caption{Performance Metrics, BLEU Score of LLMs on "CNN/Daily Mail
3.0.0" and "XSum" Datasets}
\label{table:performance-metrics-bleu}
\end{table*}

\section{Conclusion and Future Enhancements}
This research embarked on a comprehensive exploration of text summarization techniques using various Large Language Models (LLMs), with the goal of shedding light on their performance in different settings and scenarios. The study encompassed the evaluation of LLMs such as mpt-7b-instruct, falcon-7b-instruct, and text-davinci-003, as well as their summarization capabilities across two diverse datasets, 'CNN/Daily Mail 3.0.0' and 'XSum.

The experiment results, as indicated by the model performance table and human evaluation of the generated text summaries, highlight the exceptional performance of OpenAI's model, text-davinci-003, in comparison to other models. These models consistently demonstrated a superior ability to produce high-quality summaries across various datasets and temperature settings.

In the coming days, this work can be extended to leverage inferences from larger samples using higher-parameter models, such as mosaicml/mpt-30b-instruct and tiiuae/falcon-40b-instruct, potentially leading to even more robust and accurate summarization results. Additionally, the human evaluation metrics and inferences can be generated from datasets with varying word counts and output token lengths. The continual advancement of Large Language Models (LLMs) with increasing model size and capabilities offers an exciting opportunity to explore how these models can further enhance the quality of text summarization, translation, and content generation. Moreover, the fine-tuning of LLMs on specific domains and datasets could unlock the potential for domain-specific summarization models with exceptional performance.

In conclusion, this research contributes valuable insights into the field of text summarization with LLMs and offers a glimpse into future research directions. As the NLP landscape continues to evolve, leveraging the capabilities of LLMs, especially those offered by OpenAI, holds great promise for the development of advanced Generative AI applications across diverse business domains.

\section*{Acknowledgment}
The author would like to express heartfelt gratitude to Mihir Sanghvi, Mentor of the KaggleX BIPOC Mentorship Program Cohort 3. Mihir's invaluable guidance, mentorship, and insights have significantly contributed to the success of this research. Additionally, appreciation is extended to Kaggle for providing the opportunity to participate in the KaggleX BIPOC Mentorship Program, which facilitated the collaboration and learning experiences that enriched this work.

Furthermore, the support provided by Kaggle in the form of the Kaggle-KaggleX Google Cloud Platform (GCP) Coupon is acknowledged. This support enabled access to essential computing resources on Google Cloud, which was instrumental in conducting the experiments for this research. Gratitude is expressed for the collective efforts of the Kaggle community, which continues to foster a collaborative and innovative environment for data science and machine learning research.

\bibliographystyle{IEEEtran}

\end{document}